\documentclass{article}
\usepackage{graphicx}
\graphicspath{ {./images/} }

   \usepackage[final,nonatbib]{neurips_2019}

\usepackage[utf8]{inputenc} 
\usepackage[T1]{fontenc}    
\usepackage{hyperref}       
\usepackage{url}            
\usepackage{booktabs}       
\usepackage{amsfonts}       
\usepackage{nicefrac}       
\usepackage{microtype}      

\title{Raiders of the Lost Art}

\author{%
  Anthony Bourached \\
  Computer Science Department \\
  University College London \\
  London, UK \\
  \texttt{anthony.bourached.18@ucl.ac.uk} \\
  \And
  George H. Cann\thanks{www.oxia-palus.com}\\
  Oxia Palus \\
  London, UK \\
  \texttt{enquiries@oxia-palus.com} \\
}

\begin{document}

\maketitle

\begin{abstract}
  Neural style transfer, first proposed by Gatys et al. (2015), can be used to create novel artistic work through rendering a content image in the form of a style image. We present a novel method of reconstructing lost artwork, by applying neural style transfer to x-radiographs of artwork with secondary interior artwork beneath a primary exterior, so as to reconstruct lost artwork. Finally we reflect on AI art exhibitions and discuss the social, cultural, ethical, and philosophical impact of these technical innovations.
\end{abstract}



\section{Introduction}

\paragraph{}
The beauty of a piece of art is often the insight it gives the viewer into the vision of the creative mind behind it. Visual pieces are an aesthetic bridge into the artist's inspiration and perspective. This bridge does not make a one to one mapping and so conclusions are always non-unique and the meaning always remains subjective. While the primary exterior is the result with which the artist was pleased, it does not show their failed attempts; the diffuse musings of their creative process.


\begin{figure}[ht]
\centering
  \includegraphics[width=14cm]{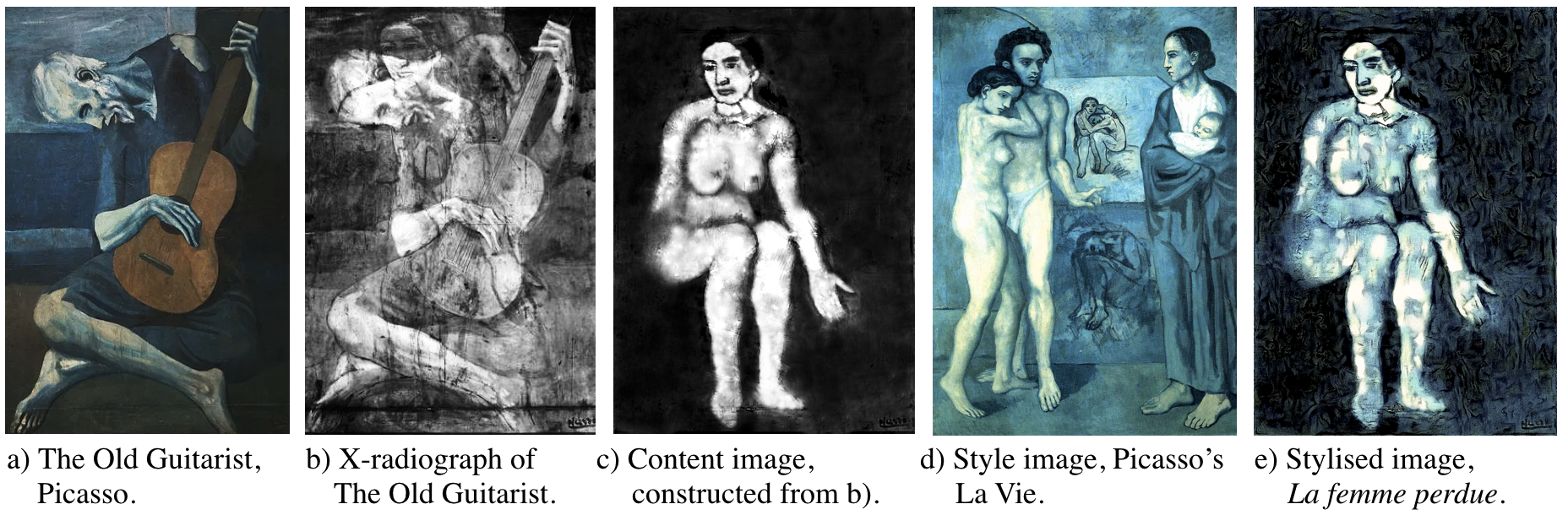}\\
  \caption{Visual method of reconstructing the art hidden beneath $\textit{The Old Guitarist}$ \cite{pablo_guitarist} \cite{pablo_la_vie}.} 
  \label{fig1}
\end{figure}

\paragraph{}
Here we attempt to broaden the insight into an artist's intentions, mistakes, and musings by reconstructing artwork that has been hidden under primary exterior of some of the most inspirational artists in history. We believe that this gives one of many possible inferences of what inspiration existed in the artist's mind. 

\section{Methods}

\paragraph{}
In this study we considered two paintings; $\textit{The Old Guitarist}$ (1903-1904), and $\textit{The Crouching Beggar}$ (1902), both painted by Pablo Picasso. Recently, art conservationists have used techniques in spectroscopy and radiography to reveal hidden art beneath a primary exterior piece. Concealed beneath $\textit{The Old Guitarist}$ hides the portrait of a woman and below $\textit{The Crouching Beggar}$ a mountainous landscape. Through x-radiography these underlying pieces have been made apparent \cite{xray}.

\paragraph{}
In this study we combined the techniques of neural style transfer \cite{gatys} and x-radiography. We edited the x-radiographed artworks to remove the impression of the primary exterior artworks. This process involved manually removing features that were deemed unlikely to be present in the interior secondary pieces. For example, $\textit{The Old Guitarist's}$ eye is highly unlikely to feature on the left shoulder of the woman in the secondary interior piece. 

\begin{figure}[ht]
\centering
  \includegraphics[width=14cm]{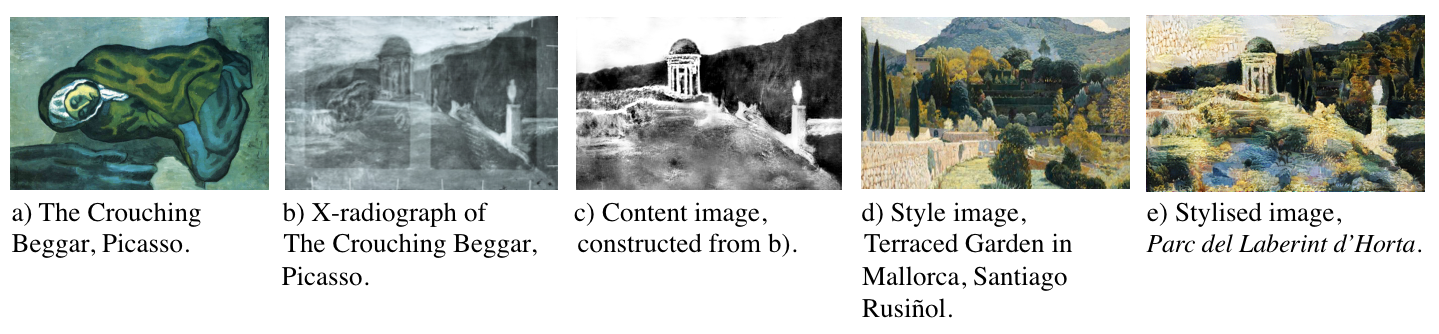}\\
  \caption{Visual method of reconstructing the artwork hidden beneath $\textit{The Crouching Beggar}$ \cite{pablo_beggar} \cite{pablo_terraced}.}\label{fig2}
\end{figure}

\paragraph{}
Using the manually edited x-radiographed image of $\textit{The Old Guitarist}$ as the content image and $\textit{La Vie}$ as the style image, we applied neural style transfer to reconstruct the lost artwork; creating the lost woman, $\textit{La femme perdue}$. $\textit{La Vie}$ was chosen as this painting was produced pre-1904, during Picasso's $\textit{Blue Period}$, and includes mutually aesthetic features. A similar process was applied to Picasso's $\textit{The Crouching Beggar}$. Beneath $\textit{The Crouching Beggar}$ lies a mountainous landscape, that art curators believe is of Parc del Laberint d’Horta, near Barcelona. They know that Santiago Rusiñol painted Parc del Laberint d’Horta. Santiago Rusiñol's $\textit{Terraced Garden in Mallorca}$ was chosen as the style image as its style and content appeared similar to that of the secondary interior piece of $\textit{The Crouching Beggar}$. Using a manually edited x-ray fluorescence image of $\textit{The Crouching Beggar}$ as the content image and Santiago Rusiñol's $\textit{Terraced Garden in Mallorca}$ as the style image, we applied neural style transfer to reconstruct the lost artwork; creating Rusiñol's \textit{Parc del Laberint d’Horta}. 

\section{Discussion}

\paragraph{}
From multiple art exhibitions, where we were the only presenters of AI art, we discovered that there is often skepticism that an algorithm can be creative, or innovate. We propose that this skepticism is rooted in the belief that a human-like, non-precise, abstract concept can not be represented in a machine. Humans' ability to empathise with something that is very similar to it, like other humans and animals, but not machines is likely the origin of this skepticism \footnote{We call this bias the \textit{Empathy Paradox}}.


\paragraph{} 
Deep convolutional neural networks enable the representation of abstract visual concepts \cite{maskrcnn}. This is largely because they combine the power of deep learning with properties such as spatial invariance which is inherent to visual patterns. The generation of images using these representations demonstrates the existence of these human-like $\textit{abstract concepts}$ in the latent space of artificial neural networks and is perhaps the only way in which non-tech experts can interface with this controversial fact. Our method of combining original but hidden artwork, subjective human input, and neural style transfer helps to broaden an insight into an artist's creative process. Furthermore, it creates a human-AI collaboration which cultivates empathy with the creative potential of AI and its harmonious use as an artistic tool.

\section{Ethical considerations}

\paragraph{}
We believe a lot of trepidation surrounding machine learning is that it replaces people. We argue that the use of machine learning as an artistic tool can ultimately broaden creative insight and widen the landscape of inspirational ingenuity by human-AI collaboration. We believe that this concept is generalisable to many domains and that it implies that the jobs of tomorrow have the opportunity to be better and more fulfilling. We believe that AI art may pioneer this positive change of mentality.

\bibliographystyle{unsrt}  
\bibliography{bibliography}

\section*{Appendix} 
\subsection*{Appendix I:}
\begin{figure}[ht]
\centering
  \includegraphics[width=14cm]{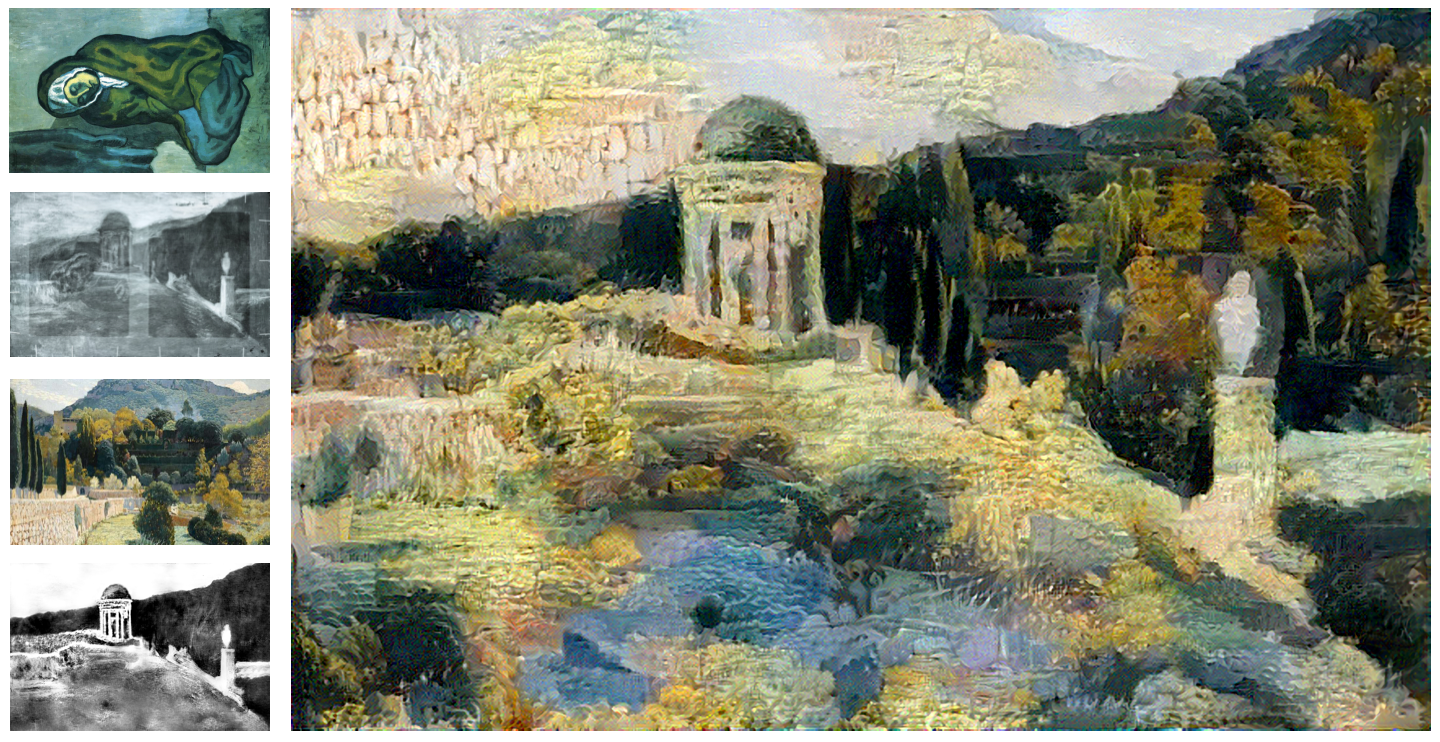}\\
  \caption{\textit{Parc del laberint d'horta}; a \textit{lost} Rusiñol, reconstructed using neural style transfer.}\label{fig3}
\end{figure}

\subsection*{Appendix II:}

\begin{figure}[ht]
\centering
  \includegraphics[width=14cm]{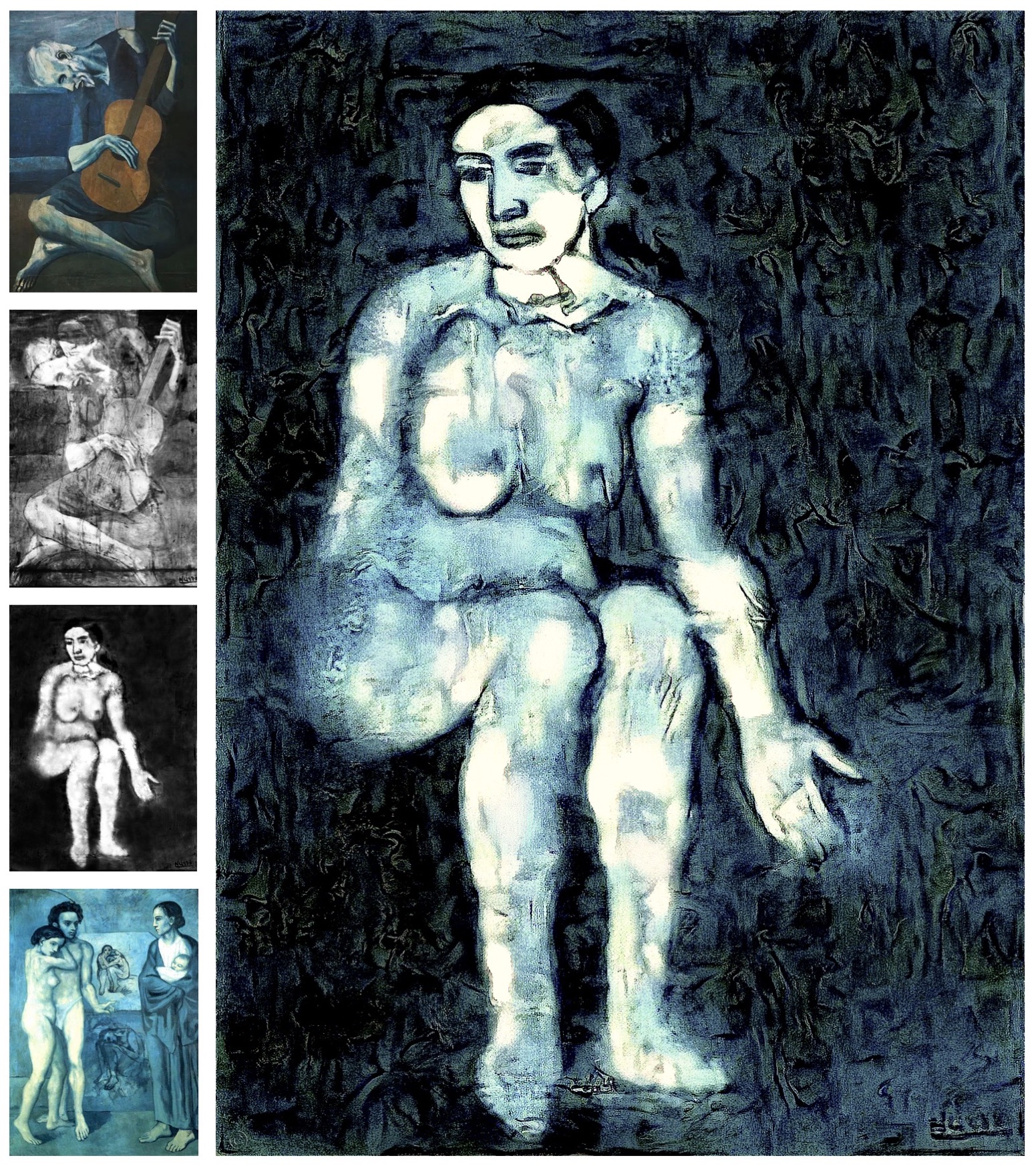}\\
  \caption{\textit{La femme perdue}; a \textit{lost} Picasso, reconstructed using neural style transfer.}\label{fig4}
\end{figure}

%
%
%
%
%

\end{document}